\documentclass[10pt,twocolumn,letterpaper]{article}

\usepackage{wacv}
\usepackage{times}
\usepackage{epsfig}
\usepackage{subcaption}
\usepackage{graphicx}
\usepackage{caption}
\usepackage{amsmath}
\usepackage{amssymb}
\usepackage{adjustbox}
\usepackage{multirow}
\usepackage{multicol}
\usepackage{booktabs}
\usepackage{url}
\usepackage{array}
\usepackage[dvipsnames,table,xcdraw]{xcolor}
\usepackage[accsupp]{axessibility}  
\usepackage{pifont}
\newcommand{\xmark}{\ding{55}}
\definecolor{myorange}{rgb}{1, 0.647, 0}
\definecolor{myblue}{rgb}{.118, 0.565, 1}
\graphicspath{Figures/}

\newcommand{\rulesep}{\unskip\hfill{\color{black}\vrule width 1pt}\hfill\ignorespaces}

\def\wacvPaperID{1068} 

\wacvalgorithmstrack   

\wacvfinalcopy 

\ifwacvfinal
\usepackage[breaklinks=true,bookmarks=false]{hyperref}
\else
\usepackage[pagebackref=true,breaklinks=true,colorlinks,bookmarks=false]{hyperref}
\fi

\begin{document}

\title{HiFormer: Hierarchical Multi-scale Representations Using Transformers for Medical Image Segmentation}

\renewcommand*{\thefootnote}{\fnsymbol{footnote}}

\author{Moein Heidari\footnotemark[1]$^{\;\, ,1}$
\quad Amirhossein Kazerouni\footnotemark[1]$^{\;\, ,1}$
\quad Milad Soltany\footnotemark[1]$^{\;\, ,1}$
\quad Reza Azad$^{2}$ \\ 
\quad Ehsan Khodapanah Aghdam$^{3}$ 
\quad Julien Cohen-Adad$^{4,5,6}$ 
\quad Dorit Merhof$\,$\footnotemark[2]$^{\;\,,7,8}$ 
\\
${^1}$ School of Electrical Engineering, Iran University of Science and Technology, Tehran, Iran\\
${^2}$ Institute of Imaging and Computer Vision, RWTH Aachen University, Aachen, Germany \\
${^3}$ Department of Electrical Engineering, Shahid Beheshti University, Tehran, Iran \\
${^4}$ Functional Neuroimaging Unit, CRIUGM, University of Montreal, Canada \\
${^5}$ NeuroPoly Lab, Institute of Biomedical Engineering, Polytechnique Montreal, Canada \\
${^6}$ MILA, Quebec AI Institute, Montreal, Canada\\
${^7}$ Faculty of Informatics and Data Science, University of Regensburg, Regensburg, Germany\\
${^8}$ Fraunhofer Institute for Digital Medicine MEVIS, Bremen, Germany\\
{\tt\small moein\_heidari@elec.iust.ac.ir, \{amirhossein477, soltany.m.99, ehsan.khpaghdam\}@gmail.com} \\ {\tt\small azad@lfb.rwth-aachen.de, jcohen@polymtl.ca, dorit.merhof@ur.de}
}
\maketitle

\footnotetext[1]{Equal contribution}
\footnotetext[2]{Corresponding author}

\thispagestyle{empty}

\begin{abstract}
Convolutional neural networks (CNNs) have been the consensus for medical image segmentation tasks. However, they suffer from the limitation in modeling long-range dependencies and spatial correlations due to the nature of convolution operation. Although transformers were first developed to address this issue, they fail to capture low-level features. In contrast, it is demonstrated that both local and global features are crucial for dense prediction, such as segmenting in challenging contexts. In this paper, we propose HiFormer, a novel method that efficiently bridges a CNN and a transformer for medical image segmentation. Specifically, we design two multi-scale feature representations using the seminal Swin Transformer module and a CNN-based encoder. To secure a fine fusion of global and local features obtained from the two aforementioned representations, we propose a Double-Level Fusion (DLF) module in the skip connection of the encoder-decoder structure. Extensive experiments on various medical image segmentation datasets demonstrate the effectiveness of HiFormer over other CNN-based, transformer-based, and hybrid methods in terms of computational complexity, quantitative and qualitative results. Our code is publicly available at~\href{https://github.com/amirhossein-kz/HiFormer}{\textcolor{magenta}{GitHub}}.
\end{abstract}

\vspace{-1.5em}
\section{Introduction} Medical image segmentation is one of the main challenges in computer vision, which provides valuable information about the areas of anatomy needed for a detailed analysis. This information can greatly assist doctors in depicting injuries, monitoring disease progression, and assessing the need for appropriate treatment. As a result of the growing use of medical image analysis, highly precise and robust segmentation has become increasingly vital. 

With their impressive ability to extract image features, Convolutional Neural Networks (CNNs) have been used widely for different image segmentation tasks. With the rise of encoder-decoder-based networks, like Fully Convolutional Networks (FCNs) \cite{long2015fully}, U-shaped structures, e.g. U-Net \cite{ronneberger2015u}, and their variants, CNNs have experienced remarkable success in medical image segmentation tasks. In both structures, skip connections are employed to embody high-level and fine-grained features provided by the encoder and decoder paths, respectively. Despite the success of CNN models in various computer vision tasks, their performance is restricted due to their limited receptive field and the inherent inductive bias \cite{dosovitskiy2020image,azad2021texture}. The aforementioned reasons prevent CNNs from building global contexts and long-range dependencies in images and, therefore, capping their performance in image segmentation.

Recently, motivated by the outstanding success of transformers in Natural Language Processing (NLP) \cite{vaswani2017attention}, vision transformers have been developed to mitigate the drawbacks of CNNs in image recognition tasks \cite{dosovitskiy2020image}. Transformers primarily leverage a multi-head self-attention (MSA) mechanism that can effectively construct long-range dependencies between the sequence of tokens and capture global contexts. The vanilla vision transformer \cite{dosovitskiy2020image} exhibits comparable performance with CNN-based methods but requires large amounts of data to generalize and suffers from quadratic complexity. Several approaches have been proposed to address these limitations. DeiT \cite{touvron2021training} proposes an efficient knowledge distillation training scheme to overcome the difficulty of vision transformers demanding a great deal of data to learn. Swin Transformer \cite{liu2021swin} and pyramid vision transformer \cite{wang2021pyramid} attempt to reduce vision transformers' computational complexity by exploiting window-based and spatial reduction attention, respectively. 

Moreover, multi-scale feature representations have lately demonstrated powerful performance in vision transformers. CrossViT \cite{chen2021crossvit} proposes a novel dual-branch transformer architecture that extracts multi-scale contextual information and provides more fine-grained feature representations for image classification. Similarly, DS-TransUNet \cite{lin2022ds} presents a dual-branch Swin Transformer to capture different semantic scale information in the encoder for the task of medical image segmentation. 
HRViT \cite{gu2022multi} connects multi-branch high-resolution architectures with vision transformers for semantic segmentation. As a result, such structures can effectively aid in enhancing the modeling of long-range relationships between tokens and obtaining more detailed information. 

Despite the vision transformers' ability to model the global contextual representation, the self-attention mechanism induces missing low-level features. Hybrid CNN-transformer approaches have been proposed to alleviate the problem above by leveraging the locality of CNNs and the long-range dependency character of transformers to encode both global and local features, particularly TransUnet \cite{chen2021transunet} and LeVit-Unet \cite{xu2021levit} in medical image segmentation. However, these approaches have some impediments that prevent them from attaining higher performance: 1) they cannot effectively combine low-level and high-level features while maintaining feature consistency, and 2) they do not use the multi-scale information produced by the hierarchical encoder properly. 

In this paper, we propose a novel encoder-decoder CNN-transformer-based framework that efficiently leverages the global long-range relationships of transformers and local feature representations of CNNs for an accurate medical image segmentation task. The encoder comprises three modules: two hierarchical CNN and Swin Transformer modules and the DLF module. Swin Transformer and CNN modules each contain three levels. First, an input image is fed into a CNN module to learn its local semantic representation. To compensate for the lack of global representation, the Swin Transformer module is applied on top of CNN's shallow features to capture long-range dependencies. Next, a pyramid of Swin Transformer modules with varying window sizes is utilized to learn multi-scale interaction. To encourage feature reusability and provide localization information, a skip connection module is designed to transfer CNN's local features into the Transformer blocks. The resulting representation of the smallest and largest pyramid levels is then entered into the DLF module. The novel proposed DLF module is a multi-scale vision transformer that fuses two obtained feature maps using a cross-attention mechanism. Finally, both recalibrated feature maps are passed into the decoder block to produce the final segmentation mask. Our proposed HiFormer not only alleviates the problem mentioned above but also surpasses all its counterparts in terms of different evaluation metrics. Our main contributions: 

\noindent $\bullet$ A novel hybrid method that merges the long-range contextual interactions of the transformer and the local semantic information of CNN.

\noindent $\bullet$ A DLF module to establish effective feature fusion between coarse and fine-grained feature representations.

\noindent $\bullet$ Experimental results demonstrate the effectiveness and superiority of the proposed HiFormer compared to the competing methods on medical image segmentation datasets.

  
  

\section{Related Works}
\vspace{-0.5em}
\subsection{CNN-based Segmentation Networks}
\vspace{-0.5em}
Convolutional Neural Networks are considered the de-facto standard for different computer vision tasks. One area where CNNs have achieved excellent results is image segmentation, where class labels are assigned to each pixel. Long et al. \cite{long2015fully} showed that fully convolutional networks (FCNs) can be used to segment images without fully connected layers. Given that the output from vanilla FCNs, where the convolutional layers are stacked sequentially, is usually coarse, other models were proposed that fuse the output of different layers \cite{badrinarayanan2017segnet,noh2015learning,ronneberger2015u}. Several approaches have been introduced to improve the limited receptive field of FCN, including dilated convolution \cite{chen2014semantic,yu2015multi} and context modeling \cite{zhao2017pyramid,chen2017deeplab}. CNN models have shown outstanding performance in medical imaging tasks. After the introduction of U-net \cite{ronneberger2015u}, other researchers focused on utilizing U-shaped encoder-decoder structures. In \cite{valanarasu2020kiu}, an over-complete network is augmented with U-net, and in U-net++ \cite{zhou2018unet++}, the encoder-decoder architecture is re-designed by adding dense skip connection between the modules. This structure has been further improved and utilized in different medical domains \cite{cai2020dense,huang2020unet,feyjie2020semi,azad2022smu}.

\vspace{-0.25em}
\subsection{Vision Transformers}
\vspace{-0.5em}
Following the remarkable success of transformers in NLP \cite{vaswani2017attention}, Dosovitskiy et al. \cite{dosovitskiy2020image} propose the Vision Transformer (ViT), which achieved state-of-the-art performance on image classification tasks by employing self-attention mechanisms to learn global information. Several derivatives of vision transformers have been introduced to make them more efficient and less dependent on a large-sized dataset to achieve generalization \cite{touvron2021training,yuan2021tokens,xia2022vision}. 


In addition, many approaches have been presented, focusing on multi-scale representations to improve accuracy and efficiency via extracting information from different scales. Inspired by the pyramid structure in CNNs \cite{newell2016stacked,yang2015multi,cai2016unified,lin2017feature}, PVT \cite{wang2021pyramid} was the first introduced pyramid vision transformer. Later, Swin Transformer \cite{liu2021swin} proposes a hierarchical vision transformer using an efficient shifted windowing approach for computing self-attention locally. CrossViT \cite{chen2021crossvit} suggests using a dual-branch vision transformer followed by a cross-attention module for richer feature representations while performing in linear time. Vision transformers have also shown impressive results in other vision tasks, including \cite{zhu2020deformable,fang2021you}, which offer end-to-end transformer-based models for object detection, and \cite{strudel2021segmenter,guo2021sotr} for semantic and instance segmentation.
\vspace{-0.25em}
\subsection{Transformers for Medical Image Segmentation}
\vspace{-0.5em}
Despite the encouraging results of CNN models, such approaches generally demonstrate restrictions for modeling long-range dependencies due to their limited receptive field, thereby yielding weak performance. Recently, transformer-based models have gained significant popularity over CNN models in medical image segmentation. Swin-UNet \cite{cao2021swin} and DS-TransUNet \cite{lin2022ds} propose pure transformer models with a U-shaped architecture based on Swin Transformer for 2D segmentation. 
In addition to fully transformer models, TransUNet \cite{chen2021transunet} takes advantage of both CNNs and transformers to capture both low-level and high-level features. UNETR \cite{hatamizadeh2022unetr} uses a transformer-based encoder to embed input 3D patches and a CNN-based decoder to achieve the final 3D segmentation results. Most prior works utilize either CNN, lacking in global features, or transformers, limited in local feature representation for feature extraction; this renders ineffective feature maps that do not contain rich information. In hybrid works, simple feature-fusing mechanisms are employed that cannot guarantee feature consistency between different scales. Motivated by multi-scale representations, we propose HiFormer, a CNN-transformer-based architecture that effectively incorporates both global and local information and utilizes a novel transformer-based fusing scheme to maintain feature richness and consistency for the task of 2D medical image segmentation.

\begin{figure*}[ht]
\centering
\begin{subfigure}{0.7\textwidth}
    \includegraphics[width=\textwidth]{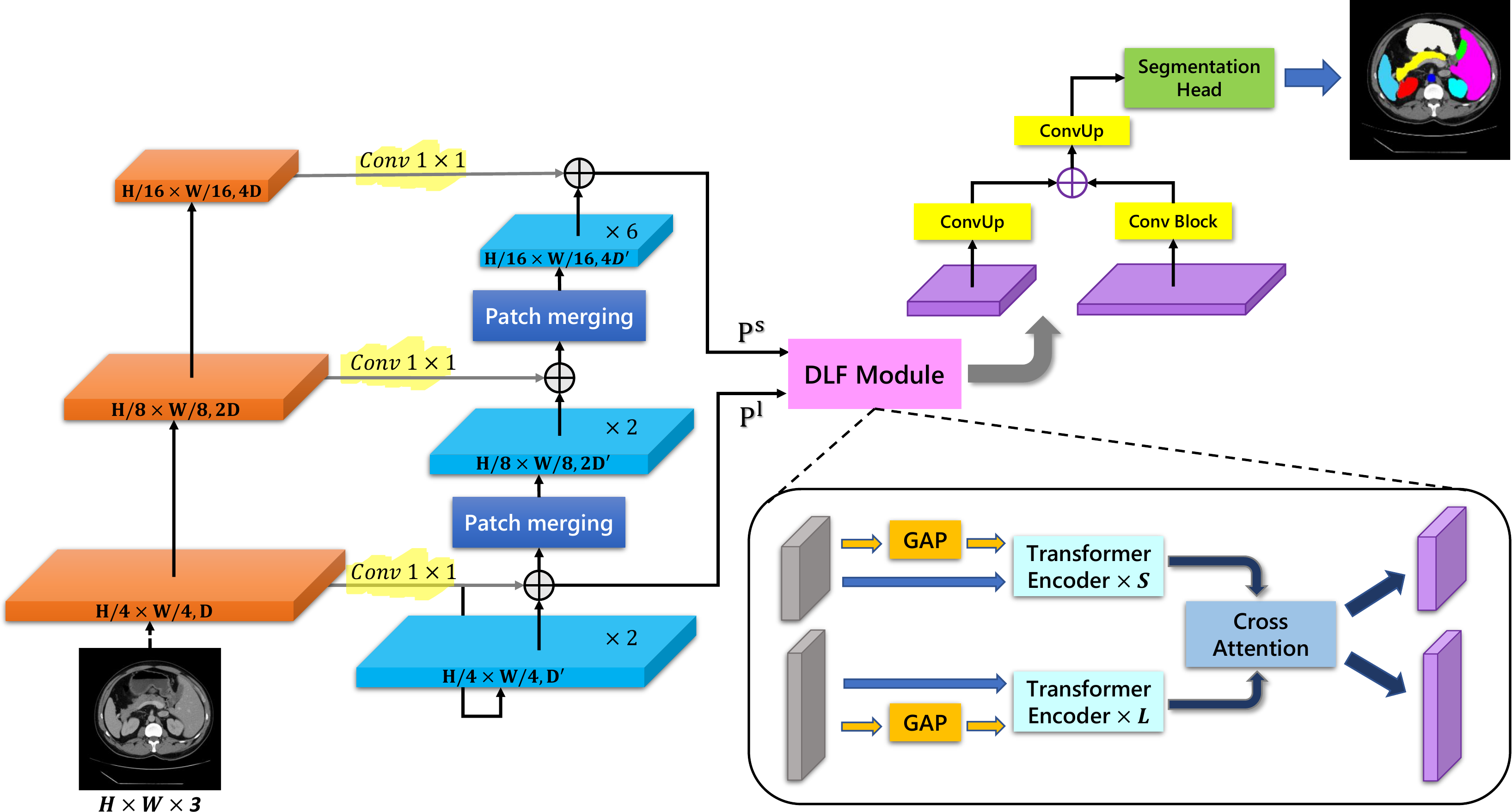}
    \caption{}
    \label{fig:Model}
\end{subfigure}
\rulesep
\begin{subfigure}{0.25\textwidth}
    \includegraphics[width=\textwidth]{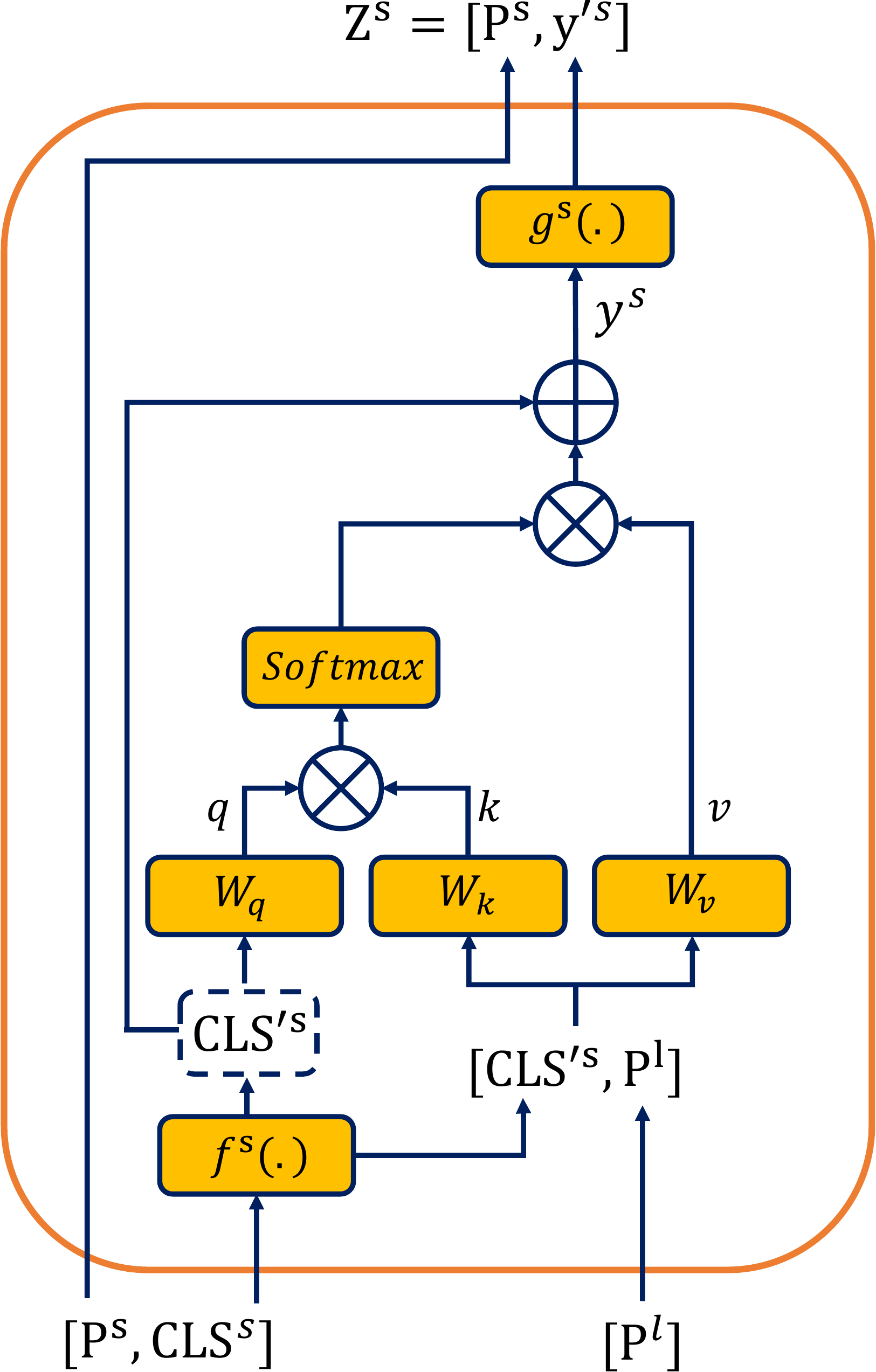}
    \caption{}
    \label{fig:MCA}
\end{subfigure}
\vspace{-0.75em}
\caption{\textbf{(a) The overview of the proposed HiFormer.} HiFormer consists of a hierarchical CNN-transformer feature extractor module; outputs of the first and last levels are fed through the DLF feature fusion module. Afterward, the decoder uses the DLF's output to generate accurate segmentation maps. In the diagram, \textcolor{myblue}{blue} and \textcolor{myorange}{orange} blocks denote Swin Transformer and CNN levels, respectively. \textbf{(b) The overview of Cross Attention.} The class token of the small level, $CLS^s$, is first projected for dimension alignment and then appended to $P^l$. The resulting embedding performs as a key and value. Moreover, $CLS'^s$ is utilized for the query. Finally, after computing attention and back projection, $Z^s$ is obtained.  This process can also be extended to the large level.}
\vspace{-1.5em}
\label{fig:mainfig}
\end{figure*}

\vspace{-0.5em}
\section{Method}
\vspace{-0.5em}
An overview of the proposed HiFormer is presented in this section. As illustrated in Fig. \ref{fig:Model}, our proposed architecture provides an end-to-end training strategy that integrates global contextual representations from Swin Transformer and local representative features from the CNN module in the encoder. A richer feature representation is then obtained using the Double-level Fusion module (DLF). Afterward, the decoder outputs the final segmentation map.

\subsection{Encoder}
\vspace{-0.5em}
As shown in Fig. \ref{fig:Model}, the proposed encoder is composed of two hierarchical models, CNN and Swin Transformer, with the DLF module that enriches the retrieved features and prepares them to be fed into the decoder. Since using CNNs or transformers separately causes either local or global features to be neglected, which affects the model's performance, we first utilize the CNN locality trait to obtain local features. Here, the CNN and Swin Transformer each include three distinct levels. We transfer local features of each level to the corresponding Swin Transformer's level via a skip connection to attain universal representations. Then each transferred CNN level is added with its parallel transformer level and passes through a Patch Merging module to produce a hierarchical representation (see Fig.~\ref{fig:Model}). We exploit the hierarchical design to take advantage of multi-scale representations. The largest and smallest levels go into the DLF module to exchange information from different scales and generate more powerful features. In the following, we will discuss our CNN, Swin Transformer, and DLF modules deeply and in detail.
\vspace{-1.5em}
\subsubsection{CNN Module}
\vspace{-0.75em}
The proposed encoder begins by employing a CNN as the feature extractor to build a pyramid of intermediate CNN feature maps of different resolutions. Taking an input image $X \in R^{H\times W \times C}$ with spatial dimensions $H$ and $W$, and $C$ channels, it is first fed into the CNN module. CNN module consists of three levels, from which a skip connection is connected to the associated transformer's level using a Conv $1\times 1$ to compensate for low-level missing information of transformers and recover localized spatial information.
\vspace{-1.5em}
\subsubsection{Swin Transformer Module}
\vspace{-0.75em}
The vanilla transformer encoder block \cite{dosovitskiy2020image} consists of two main modules: a multi-head self-attention (MSA) and a multi-layer perceptron (MLP). The vanilla transformer is composed of N identical transformer encoder blocks. In each block, before the MSA and the MLP blocks, LayerNorm (LN) is applied. Additionally, a copy of the activations is also added to the output of the MSA or MLP block through skip-connections. One major problem with the vanilla ViT, which uses the standard MSA, is its quadratic complexity, rendering it inefficient for high-resolution computer vision tasks like image segmentation. To overcome this limitation, Swin Transformer \cite{cao2021swin} introduced the W-MSA and SW-MSA.

The Swin Transformer module includes two successive modified transformer blocks; the MSA block is replaced with the window-based multi-head self-attention (W-MSA) and the shifted window-based multi-head self-attention (SW-MSA). In the W-MSA module, self-attention will be applied to local windows of size $M \times M$. The W-MSA module has linear complexity; however, given that there is no connection across windows, it has limited modeling power. To alleviate this, SW-MSA is introduced that utilizes a windowing configuration that is shifted compared to the input of the W-MSA module; this is to ensure that we have cross-window connections. This process is depicted in Eq \ref{eq.swin}. 
\vspace{-0.5em}
\begin{align}
    &{{\hat{\bf{z}}}^{l}} = \text{W-MSA}\left( {\text{LN}\left( {{{\bf{z}}^{l - 1}}} \right)} \right) + {\bf{z}}^{l - 1},\nonumber\\
    &{{\bf{z}}^l} = \text{MLP}\left( {\text{LN}\left( {{{\hat{\bf{z}}}^{l}}} \right)} \right) + {{\hat{\bf{z}}}^{l}},\nonumber\\
    &{{\hat{\bf{z}}}^{l+1}} = \text{SW-MSA}\left( {\text{LN}\left( {{{\bf{z}}^{l}}} \right)} \right) + {\bf{z}}^{l}, \nonumber\\
    &{{\bf{z}}^{l+1}} = \text{MLP}\left( {\text{LN}\left( {{{\hat{\bf{z}}}^{l+1}}} \right)} \right) + {{\hat{\bf{z}}}^{l+1}}, \label{eq.swin}
    \vspace{-1em}
\end{align}
The output of the first level in the CNN pyramid will be fed into a $1\times1$ convolution to generate $(H/4 \times W/4)$ patches (tokens) of length $D'$. These patches go through the first Swin Transformer block, generating the first attention-based feature maps. A skip-connection adds the previous activations to the obtained feature maps, resulting in the largest branch feature map $P^l$. Next, patch-merging is applied, which concatenates $2\times2$ groups of adjacent patches, applies a linear layer, and increases the embedding dimensions from $D'$ to $2D'$ while reducing resolution. Similarly, higher-level feature maps of both the CNN and attention-based feature maps are fused and fed into Swin Transformer blocks to generate higher-level outputs. The latter is denoted as $P^s$, the smallest level feature map.
\vspace{-0.5cm}
\subsubsection{Double-Level Fusion Module (DLF)}
\vspace{-0.5em}
The main challenge is efficiently fusing CNN and Swin Transformer level features while preserving feature consistency. A straightforward approach is to directly feed the summation of CNN levels with their matching Swin Transformer levels through a decoder and attain the segmentation map. Such an approach, however, fails to ensure feature consistency between them, leading to subpar performance. Hence, we propose a novel Double-Level Fusion (DLF) module, which takes the resultant smallest ($P^{s}$) and largest ($P^{l}$) levels as inputs and employs a cross-attention mechanism to fuse information across scales. 

In general, shallow levels have better localization information, and as we approach deeper levels, semantic information becomes more prevalent and is better suited for the decoder part. Faced with the dilemma of extensive computational cost and the imperceptible effect of the middle-level feature map in model accuracy, we did not consider using the middle level in feature fusion to save computational costs. As a result, we encourage representation by multiscaling the shallowest ($P^s$) and last ($P^l$) levels while preserving localization information.

In the proposed DLF module, the class token plays a significant role since it summarizes all the information of input features. We assign each level a class token derived from global average pooling (GAP) over the level's norm. We obtain class tokens as demonstrated below:

\begin{equation}
    \vspace{-0.25em}
    \begin{aligned}
        CLS^{s} &= GAP(Norm (P^{s}))
        \\
        CLS^{l} &= GAP(Norm (P^{l}))
    \end{aligned}
\end{equation}
where $CLS^{s}$ $\in R^{4D' \times 1}$ and $CLS^{l}$ $\in R^{D'\times 1}$. Class tokens are then concatenated with associated level embeddings before being passed into the transformer encoders. The small level is followed by $S$ and the large level by $L$ transformer encoders for computing global self-attention. Notably, we also add a learnable position embedding for each token of both levels before giving them to the transformer encoders for learning position information.

After passing embeddings through the transformer encoders, features of each level are fused using the cross-attention module. Specifically, before fusion, two-level class tokens are swapped, which means the class token of one level concatenates with the tokens of the another level. Then each new embedding is separately fed through the module for fusion and finally back-projected to its own level. This interaction with other level tokens enables class tokens to share rich information with their cross-level. 

In particular, this displacement for the small level is shown in Fig. \ref{fig:MCA}. $f^s(.)$ first projects $CLS^s$ to the dimensionality of $P^l$, and the output is denoted as $CLS'^s$. $CLS'^s$ concatenated with $P^l$ serves as a key and value and independently performs as a query for computing attention. Since we only query the class token, the cross-attention mechanism operates in linear time. The final output $Z^s$ can be mathematically written as follows:
\begin{align}
\vspace{-3em}
    &y^{s} = f^{s}(CLS^s)+MCA(LN([f^s(CLS^s)\parallel P^l])) 
    \nonumber\\
    &Z^s=[P^s \parallel g^s(y^s)] 
    \label{eq.MCA}
    \vspace{-1.5em}
\end{align}
 
\vspace{-1.2em}
\subsection{Decoder}
\vspace{-0.5em}
Motivated by Semantic FPN \cite{kirillov2019panoptic}, we design a decoder that combines features from the $P^{s}$ and $P^{l}$ levels into a unified mask feature.
First, the low and high-resolution feature maps, $P^{s}$ and $P^{l}$, are received from the DLF module. $P^{s}$ (H/16, W/16) is followed by a ConvUp block which applies two stages of $3\times3$ Conv, $2\times$ bilinear upsampling, Group Norm \cite{wu2018group}, and ReLU to attain ( H/4, W/4 ) resolution. $P^{l}$ ( H/4, W/4 ) is also followed by a Conv Block, which employs a $3\times3$ Conv, Group Norm, and ReLU and remains at~( H/4, W/4 ) resolution. The summation of both processed $P^{s}$ and $P^{l}$ is headed through another ConvUp block to achieve the final unified $H\times W$ feature map. After passing the acquired feature map through a $3\times3$ Conv in the segmentation head, the final segmentation map is generated.

\vspace{-0.5em}
\section{Experiments}
\vspace{-0.5em}
\subsection{Dataset}
\vspace{-0.5em}
\noindent\textbf{Synapse Multi-Organ Segmentation:}
First, we evaluate HiFormer's performance on the benchmarked synapse multi-organ segmentation dataset \cite{synapse2015ct}. This dataset includes 30 cases with 3779 axial abdominal clinical CT images where each CT volume involves $85 \sim 198$ slices of $512 \times 512$ pixels, with a voxel spatial resolution of $([0.54 \sim 0.54] \times[0.98 \sim 0.98] \times[2.5 \sim 5.0])$ $\mathrm{mm}^{3}$. 

\noindent\textbf{Skin Lesion Segmentation:}
We conduct extensive experiments on the skin lesion segmentation datasets. Specifically, we utilize the ISIC 2017 dataset \cite{codella2018skin} comprising 2000 dermoscopic images for training, 150 for validation, and 600 for testing. Moreover, we adopt the ISIC 2018 \cite{codella2019skin} and follow the literature work \cite{asadi2020multi,azad2019bi} to divide the dataset into the train, validation, and test sets accordingly. Besides, the $\mathrm{PH}^{2}$ dataset \cite{mendoncca2013ph} is used, a dermoscopic image database introduced for both segmentation and classification tasks. 

\noindent\textbf{Multiple Mylomia Segmentation:}
We also evaluate our methodology on multiple myeloma cell segmentation grand challenges provided by SegPC 2021 \cite{segpc2021,gupta2018pcseg}. The challenge dataset includes a training set with 290 samples and validation and test sets with 200 and 277 samples, respectively. 
\vspace{-1.5em}
\subsection{Implementation Details} 
\vspace{-0.5em}
We implemented our framework in PyTorch and trained on a single Nvidia RTX 3090 GPU with 24 GB of memory. The input image size is $224\times224$, and we set the batch size and learning rate to 10 and 0.01 during training, respectively. In addition, we use the weights pre-trained on ImageNet for the CNN and Swin Transformer modules to initialize their parameters. Our model is optimized using the SGD optimizer with a momentum of 0.9 and weight decay of 0.0001. Moreover, data augmentations such as flipping and rotating are employed during training to boost diversity. Table \ref{table:configurations} depicts the suggested model's final configurations.
\begin{table}[ht]
    \centering
    \vspace{-0.5em}
    \caption{\textbf{The proposed model configurations.} WS represents window size, $D'$ expresses the embedding dimension, and $r$ denotes the MLP expanding ratio used in the transformer block. The number of heads in the DLF module is the same for both levels.}
    \vspace{-0.5em}
    \label{table:configurations}
    \resizebox{\columnwidth}{!}{%
    \begin{tabular}{c|c|cccc|cccccc}
    \toprule
    \multirow{3}{*}{\textbf{Model}} & \multirow{3}{*}{\textbf{CNN}} & \multicolumn{4}{c|}{\multirow{2}{*}{\textbf{Swin Transformer}}} & \multicolumn{6}{c}{\textbf{DLF}} \\
               &          & \multicolumn{4}{c|}{} & \multicolumn{2}{c}{Dimension} &   &   &   &         \\
               &          & $D'$                  & \# Layer        & \# Head          & WS & $\mathrm{P^s}$        & $\mathrm{P^l}$        & S & L & r & \# Head 
    \\ \midrule
    HiFormer-S & ResNet34 & 96  & [2,2,6] & [3,6,12] & 7  & 384  & 96  & 1 & 1 & 1 & 3       \\
    HiFormer-B & ResNet50 & 96  & [2,2,6] & [3,6,12] & 7  & 384  & 96  & 2 & 1 & 2 & 6   \\
    HiFormer-L & ResNet34 & 96  & [2,2,6] & [3,6,12] & 7  & 384  & 96  & 4 & 1 & 4 & 6      
    \\ \bottomrule
    \end{tabular}%
    }
    \vspace{-0.5em}
\end{table}

\begin{figure*}[ht]
	\centering
	\includegraphics[width=0.95\textwidth]{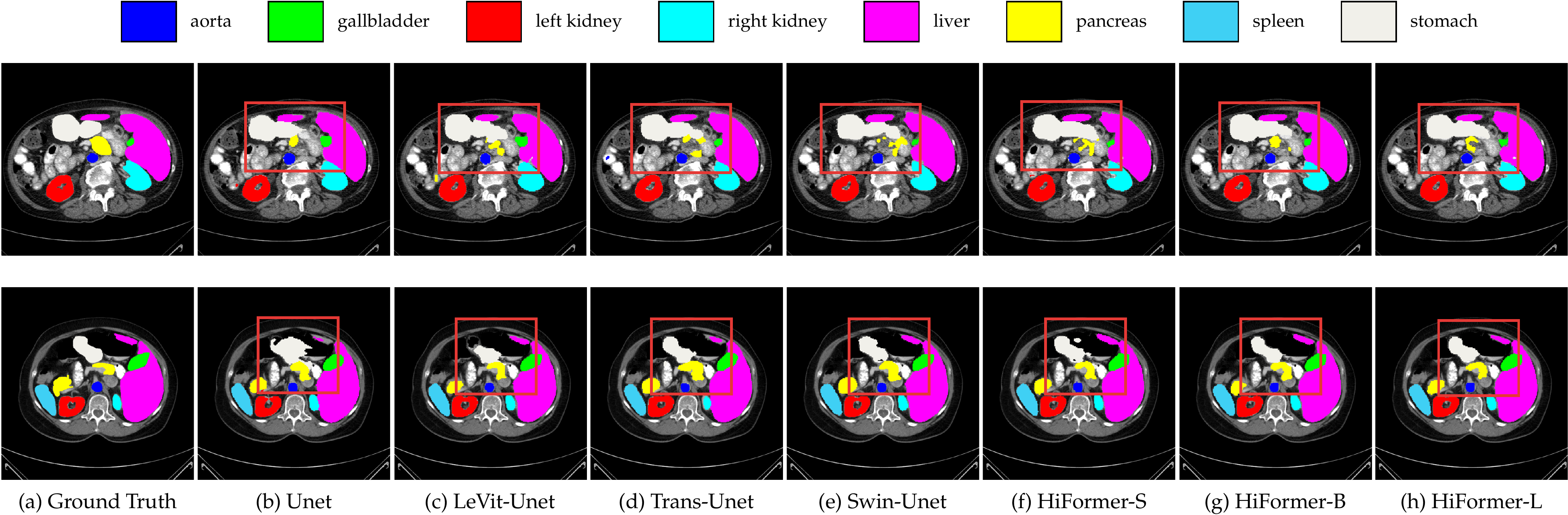}
 \vspace{-1em}
	\caption{Segmentation results of the proposed method on the \textit{Synapse} dataset. The red rectangles identify organ regions where the superiority of our proposed method can be clearly seen.}
	\label{fig:synapseviz}
	\vspace{-1.5em}
\end{figure*}
\vspace{-0.75em}
\subsection{Evaluation Results}
\vspace{-0.5em}
We adopt a task-specific paradigm in terms of evaluation metrics in each experiment. Specifically, these metrics include the Dice score, 95\% Hausdorff Distance (HD), Sensitivity and Specificity, Accuracy, and mIOU. To ensure an unprejudiced comparison, we contrast HiFormer against both CNN and transformer-based methods, along with the models formulated on an amalgamation of both.
\vspace{-1em}
\subsubsection{Results of Synapse Multi-Organ Segmentation}
\vspace{-0.75em}
The comparison of the proposal with previous state-of-the-art (SOTA) methods in terms of the average Dice-Similarity Coefficient (DSC) and average Hausdorff Distance (HD) on eight abdominal organs is shown in Table~\ref{comparison}. HiFormer outperforms CNN-based SOTA methods by a large margin. Compared to other transformer-based models, our HiFormer-B shows superior learning ability on both evaluation metrics, observing an increase of $2.91\%$ and $1.26\%$ in Dice score and a decrease of $16.99$ and $6.85$ in average HD compared to TransUnet and Swin-Unet, respectively. Concretely, HiFormer steadily beats the literature work in the segmentation of most organs, particularly for the stomach, kidney, and liver segmentation. One can observe that HiFormer has distinct advantages over other methods in terms of average HD. Besides, the efficiency in terms of the number of parameters is indicated in Table~\ref{comparison}, which will be discussed in the following sections. A characteristic qualitative example of the results is given in Fig.~\ref{fig:synapseviz}. We have observed that the proposed method can accurately segment fine and complex structures and output more accurate segmentation results, which are more robust to complicated backgrounds.
\vspace{-2.5em}
\subsubsection{Results of Skin Lesion Segmentation}
\begin{figure*}[htb]
    \centering
    \resizebox{0.92\textwidth}{!}{
    \begin{tabular}{@{} *{7}c @{}}
    \includegraphics[width=0.16\textwidth]{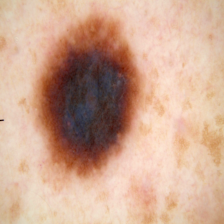} &
    \includegraphics[width=0.16\textwidth]{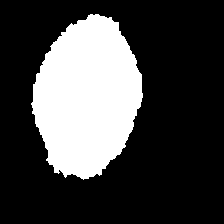} &
    \includegraphics[width=0.16\textwidth]{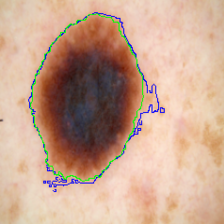} &
    \includegraphics[width=0.16\textwidth]{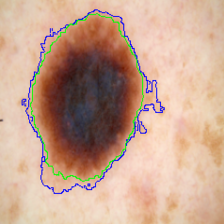} &
    \includegraphics[width=0.16\textwidth]{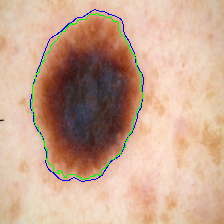} &
    \includegraphics[width=0.16\textwidth]{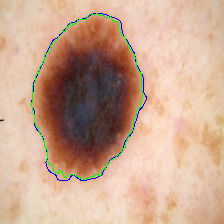} &
    \includegraphics[width=0.16\textwidth]{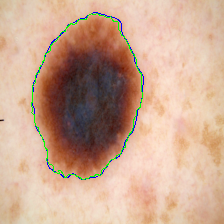} \\
    \includegraphics[width=0.16\textwidth]{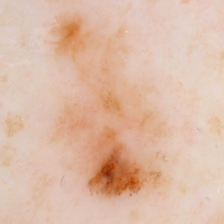} &
    \includegraphics[width=0.16\textwidth]{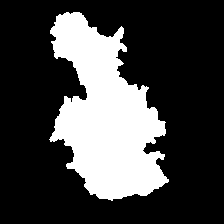} &
    \includegraphics[width=0.16\textwidth]{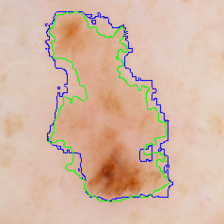} &
    \includegraphics[width=0.16\textwidth]{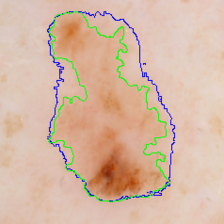} &
    \includegraphics[width=0.16\textwidth]{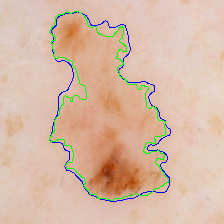} &
    \includegraphics[width=0.16\textwidth]{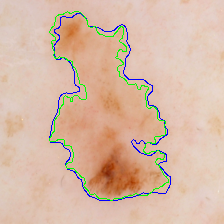} &
    \includegraphics[width=0.16\textwidth]{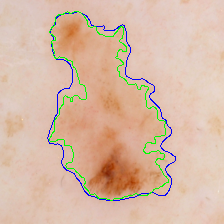} \\
    { \large (a) Input Image} & {\large (b) Ground Truth} & {\large (c) Swin-Unet} & {\large (d) TMU-Net} & {\large (e) HiFormer-S} & {\large (f) HiFormer-B} & {\large (g) HiFormer-L}
    \end{tabular}
}
    \vspace{-0.75em}
    \caption{Visual comparisons of different methods on the \textit{ISIC2017} skin lesion segmentation dataset. Ground truth boundaries are shown in \textcolor{green}{green}, and predicted boundaries are shown in \textcolor{blue}{blue}.} 
    \label{fig:skin}
    \vspace{-1em}
\end{figure*}

\vspace{-0.75em}
The comparison results for benchmarks of ISIC 2017, ISIC 2018, and $\mathrm{PH}^{2}$ skin lesion segmentation task against leading methods are presented in Table~\ref{tab:skin comparison}. Our HiFormer performs much better than other competitors w.r.t. most of the evaluation metrics. Specifically, the superiority of HiFormer across different datasets highlights its satisfactory generalization ability. 
We also show a visual comparison of the skin lesion segmentation results in Fig.~\ref{fig:skin} which indicates that our proposed method is able to capture finer structures and generates more precise contours. Specifically, as in Fig. \ref{fig:skin}, our approach performs better than hybrid methods such as TMU-Net \cite{reza2022contextual} in boundary areas. Moreover, showcased in Fig. \ref{fig:skin}, HiFormer is robust to noisy items compared to pure transformer-based methods such as Swin-Unet \cite{cao2021swin}, where the performance degrades due to lack of locality modeling. The superior performance is achieved by an expedient combination of transformer and CNN for modeling global relationships and local representations.

\begin{table*}[!ht]
    \centering
    \caption{Comparison results of the proposed method on the \textit{Synapse} dataset. \textcolor{blue}{Blue} indicates the best result, and \textcolor{red}{red} displays the~second-best.}
    \vspace{-0.5em}
    \label{comparison}
    \resizebox{\textwidth}{!}{
    \begin{tabular}{l|cc|cccccccc} 
    \toprule
    \textbf{Methods} & \textbf{DSC~$\uparrow$} & \textbf{HD~$\downarrow$} & \textbf{Aorta} & \textbf{Gallbladder} & \textbf{Kidney(L)} & \textbf{Kidney(R)} & \textbf{Liver} & \textbf{Pancreas} & \textbf{Spleen} & \textbf{Stomach} \\ 
    \midrule
    DARR \cite{fu2020domain} & 69.77 & - & 74.74 & 53.77 & 72.31 & 73.24 & 94.08 & 54.18 & 89.90 & 45.96 
    \\ 
    R50 U-Net \cite{chen2021transunet} & 74.68 & 36.87 & 87.74 & 63.66 & 80.60 & 78.19 & 93.74 & 56.90 & 85.87 & 74.16 
    \\
    U-Net \cite{ronneberger2015u} & 76.85 & 39.70 & \textcolor{red}{89.07} & \textcolor{red}{69.72} & 77.77 & 68.60 & 93.43 & 53.98 & 86.67 & 75.58 
    \\
    R50 Att-UNet \cite{chen2021transunet}  & 75.57 & 36.97 & 55.92 & 63.91 & 79.20 & 72.71 & 93.56 & 49.37 & 87.19 & 74.95 
    \\
    Att-UNet \cite{schlemper2019attention}  & 77.77 & 36.02 & \textcolor{blue}{89.55} & 68.88 & 77.98 & 71.11 & 93.57 & 58.04 & 87.30 & 75.75 
    \\
    R50 ViT \cite{chen2021transunet}  & 71.29 & 32.87 & 73.73 & 55.13 & 75.80 & 72.20 & 91.51 & 45.99 & 81.99 & 73.95 
    \\
    TransUnet \cite{chen2021transunet} & 77.48 & 31.69 & 87.23 & 63.13 & 81.87 & 77.02 & 94.08 & 55.86 & 85.08 & 75.62 
    \\
    Swin-Unet \cite{cao2021swin} & 79.13 & 21.55 & 85.47 & 66.53 & 83.28 & 79.61 & \textcolor{red}{94.29} & 56.58 & 90.66 & 76.60 
    \\
    LeVit-Unet \cite{xu2021levit} & 78.53 & \textcolor{red}{16.84} & 78.53 & 62.23 & \textcolor{red}{84.61} & \textcolor{blue}{80.25} & 93.11 & 59.07 & 88.86 & 72.76 
    \\
    DeepLabv3+ (CNN) \cite{chen2018encoder} & 77.63 & 39.95 & 88.04 & 66.51 & 82.76 & 74.21 & 91.23 & 58.32 & 87.43 & 73.53 
    \\
    \midrule
    \rowcolor[HTML]{C8FFFD}  
    \textbf{HiFormer-S} & 80.29 & 18.85 & 85.63 & \textcolor{blue}{73.29} & 82.39 & 64.84 & 94.22 & \textcolor{blue}{60.84} & \textcolor{blue}{91.03} & 78.07 
    \\
    \rowcolor[HTML]{C8FFFD} 
    \textbf{HiFormer-B} & \textcolor{red}{80.39} & \textcolor{blue}{14.70} & 86.21 & 65.69 &\textcolor{blue}{85.23} & \textcolor{red}{79.77} & \textcolor{blue}{94.61} & 59.52 & \textcolor{red}{90.99} & \textcolor{red}{81.08} 
    \\
    \rowcolor[HTML]{C8FFFD} 
    \textbf{HiFormer-L} & \textcolor{blue}{80.69} & 19.14 & 87.03 & 68.61 & 84.23 & 78.37 & 94.07 & \textcolor{red}{60.77} & 90.44 & \textcolor{blue}{82.03} 
    \\
    \bottomrule
    \end{tabular}
    }
\end{table*}

\begin{table*}[!ht]
\vspace{-0.25em}
    \centering
    \caption{Performance comparison of the proposed method against the SOTA approaches on skin lesion segmentation benchmarks. \textcolor{blue}{Blue} indicates the best result, and \textcolor{red}{red} displays the second-best.}
    \vspace{-0.5em}
    \label{tab:skin comparison}
    \resizebox{\textwidth}{!}{
    \begin{tabular}{l||cccc||cccc||cccc} 
    \toprule
    \multirow{2}{*}{\textbf{Methods}} & \multicolumn{4}{c||}{\textbf{ISIC 2017}} &  \multicolumn{4}{c||}{\textbf{ISIC 2018}}  & \multicolumn{4}{c}{$\mathbf{PH^2}$} \\ 
    \cline{2-13}
     & \textbf{DSC} & \textbf{SE} & \textbf{SP} & \textbf{ACC} & \textbf{DSC} & \textbf{SE} & \textbf{SP} & \textbf{ACC} & \textbf{DSC} & \textbf{SE} & \textbf{SP} & \textbf{ACC} \\ 
    \midrule
    U-Net \cite{ronneberger2015u} & 0.8159 & 0.8172 & 0.9680 & 0.9164 & 0.8545 & 0.8800 & 0.9697 & 0.9404 & 0.8936 & 0.9125  & 0.9588  & 0.9233 
    \\
    Att-UNet \cite{schlemper2019attention} & 0.8082 & 0.7998 & 0.9776 & 0.9145 & 0.8566 & 0.8674 & \textcolor{blue}{0.9863} & 0.9376 & 0.9003 & 0.9205  & 0.9640  & 0.9276 
    \\
    DAGAN \cite{lei2020skin} & 0.8425 & 0.8363 & 0.9716 & 0.9304 & 0.8807 & 0.9072 & 0.9588 & 0.9324 & 0.9201 & 0.8320  & 0.9640  & 0.9425 
    \\
    TransUNet \cite{chen2021transunet} & 0.8123 & 0.8263 & 0.9577 & 0.9207 & 0.8499 & 0.8578 & 0.9653 & 0.9452 & 0.8840 & 0.9063  & 0.9427  & 0.9200 
    \\
    MCGU-Net \cite{asadi2020multi} & 0.8927 & 0.8502 & 0.9855 & 0.9570 & 0.8950 & 0.8480 & \textcolor{red}{0.9860}  & 0.9550 & 0.9263 & 0.8322  & 0.9714  & 0.9537 
    \\
    MedT \cite{valanarasu2021medical} & 0.8037 & 0.8064 & 0.9546 & 0.9090 & 0.8389 & 0.8252 & 0.9637 & 0.9358 & 0.9122 & 0.8472  & 0.9657  & 0.9416 
    \\
    FAT-Net \cite{wu2022fat} & 0.8500 & 0.8392 & 0.9725 & 0.9326 & 0.8903 & \textcolor{red}{0.9100} & 0.9699 & 0.9578 & 0.9440 & 0.9441  & 0.9741  & \textcolor{blue}{0.9703} 
    \\
    TMU-Net \cite{reza2022contextual} & 0.9164 & 0.9128 & 0.9789 & 0.9660 & 0.9059 & 0.9038 & 0.9746 & 0.9603 & 0.9414 & 0.9395  & 0.9756 & 0.9647 
    \\
    Swin-Unet \cite{cao2021swin} & 0.9183 & 0.9142 & 0.9798 & \textcolor{red}{0.9701} & 0.8946 & 0.9056 & 0.9798 & \textcolor{blue}{0.9645} & 0.9449 & 0.9410 & 0.9564 & \textcolor{red}{0.9678} 
    \\
    DeepLabv3+ (CNN) \cite{chen2018encoder} & 0.9162 & 0.8733 & \textcolor{blue}{0.9921} & 0.9691 & 0.8820 & 0.8560 & 0.9770 & 0.9510 & 0.9202 & 0.8818 & \textcolor{blue}{0.9832} & 0.9503 
    \\
    \midrule
    \rowcolor[HTML]{C8FFFD}
    \textbf{HiFormer-S} & \textcolor{red}{0.9238} & \textcolor{red}{0.9153} & 
    0.9832 & 0.9695 & \textcolor{red}{0.9079} & 0.8934 & 0.9801 & 0.9618 & \textcolor{red}{0.9455} & \textcolor{blue}{0.9737} & 0.9604 & 0.9646
    \\
    \rowcolor[HTML]{C8FFFD}
    \textbf{HiFormer-B} & \textcolor{blue}{0.9253} & \textcolor{blue}{0.9155} & 0.9840 & \textcolor{blue}{0.9702} & \textcolor{blue}{0.9102} & \textcolor{blue}{0.9119} & 0.9755 & \textcolor{red}{0.9621} & \textcolor{blue}{0.9460} & 0.9420 & \textcolor{red}{0.9772} & 0.9661 
    \\
    \rowcolor[HTML]{C8FFFD}
    \textbf{HiFormer-L} & 0.9225 & 0.9046 & \textcolor{red}{0.9856} & 0.9693 & 0.9053 & 0.8828 & 0.9820 & 0.9611 & 0.9451 & \textcolor{red}{0.9561} & 0.9691 & 0.9659 
    \\
    \bottomrule
    \end{tabular}
    }
    \vspace{-1.5em}
\end{table*}
\vspace{-1.25em}
\subsubsection{Results of Multiple Mylomia Segmentation}
\begin{figure}[htb]
\centering
\vspace{-0.5em}
\resizebox{0.92\columnwidth}{!}{
    \begin{tabular}{*{3}c}
    {\small (a) Input Image } & {\small (b) Ground Truth } & {\small (c) Prediction} \\
    \includegraphics[width=0.14\textwidth]{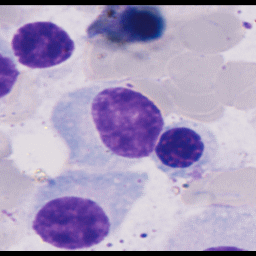} &
    \includegraphics[width=0.14\textwidth]{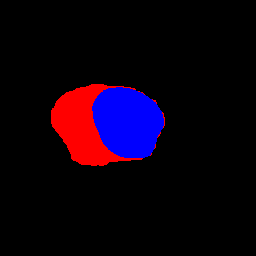} &
    \includegraphics[width=0.14\textwidth]{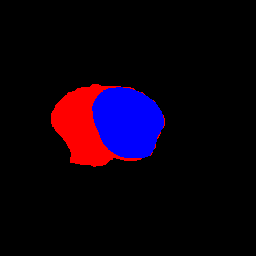} \\
    \includegraphics[width=0.14\textwidth]{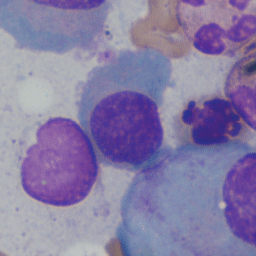} &
    \includegraphics[width=0.14\textwidth]{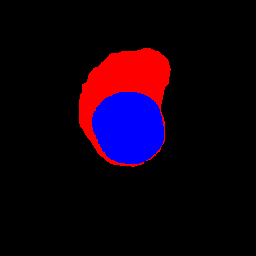} &
    \includegraphics[width=0.14\textwidth]{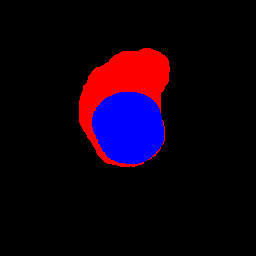} \\
    \end{tabular}
    }
    \vspace{-0.75em}
    \caption{Visual representation of the proposed method on the \textit{SegPC} cell segmentation dataset.} 
    \label{fig:segpc}
\end{figure}

\vspace{-0.5em}
In Table~\ref{tab:segPC}, we include the results based on the mean IoU metric. The HiFormer structure consistently outperformed the challenge leaderboard in all configurations we tested. In addition, some segmentation outputs of the proposed HiFormer are illustrated in Fig. \ref{fig:segpc}. As shown, our predictions adjust well to the provided GT masks. One of the key advantages of HiFormer is its ability to model multi-scale representation. It restrains the background noise, which is the case in datasets with highly overlapped backgrounds (such as SegPC). Stated succinctly, HiFormer exceeds CNN-based methods with only local information modeling ability and transformer-based counterparts, which render poor performance in boundary areas.
\vspace{-0.5em}
\begin{table}[t]
\centering
	\caption{Performance evaluation on the \textit{SegPC} challenge.}
    \vspace{-0.5em}
    \resizebox{0.8\columnwidth}{!}{
	\begin{tabular}{cc}
		\hline
		\textbf{Methods} & \textbf{mIOU}\\
		\hline
		Frequency recalibration U-Net \cite{azad2021deep} &0.9392\\		
		XLAB Insights  \cite{bozorgpour2021multi} & 0.9360 \\
		DSC-IITISM  \cite{bozorgpour2021multi} &0.9356\\
		Multi-scale attention deeplabv3+ \cite{bozorgpour2021multi} &0.9065\\
		U-Net \cite{ronneberger2015u} &0.7665\\
		Contextual attention \cite{reza2022contextual}& 0.9395\\
		\hline
		HiFormer-S& 
		0.9392\\
		\textbf{HiFormer-B} & \textbf{0.9406}\\
		HiFormer-L& 
		0.9395\\
		\hline
	\end{tabular}
	}
	\label{tab:segPC}
	\vspace{-2em}
\end{table}

\subsection{Comparison of Model Parameters}
\vspace{-0.5em}
In \ref{table:model parameters}, we compare the numbers of parameters of our proposed method with those of medical image segmentation models.
Our lightweight HiFormer shows great superiority in terms of model complexity while attaining eminent or on-par performance compared to the literature works.
\vspace{-0.5em}
\section{Ablation Study}
\vspace{-0.5em}
\noindent\textbf{Comparison of different CNN backbones.}
We first investigate the contribution of different CNN backbones. Specifically, we employ variants of ResNet \cite{he2016deep} and DenseNet \cite{huang2017densely} as two prior arts of
convolutional architectures. As shown in Table~\ref{table:CNN}, utilizing the ResNet backbone results in the best performance. Moreover, we have witnessed that a larger CNN backbone does not necessarily result in a performance boost (see rows 3 and 4 in Table~\ref{table:CNN}), which gives us the insight to use ResNet50 architecture as the default. 

\noindent\textbf{Impact of the DLF module.}
Next, we evaluate the importance of the DLF module on segmentation performance. The experimental results reported in Table~\ref{table:DLF} reveal the non-negligible role of the DLF module during the encoding and decoding process. Specifically, the DLF module brings significant improvements (3.24\% and 2.18\%) to the dice score and HD, respectively. Through the cross-attention mechanism, the DLF module assists the network in incorporating global and local features. The results prove that an expedient combination of CNN and transformer is helpful in segmenting target lesions. In addition, the impact of the DLF module on the SegPc and Skin datasets are provided in the Supplementary Material (SM) (see Table 1-2).

\noindent\textbf{Ablation on different DLF module configurations.}
Table~\ref{table:DLF Configs} shows the performance of different DLF module configurations. We test different values for the number of heads and depth (S and L) for both the small and large levels and the MLP expanding ratio in the MLP block of the transformer module (r). We observe that the pair of $(2,1)$ for $(S, L)$ and six heads for both levels work best. As shown in row A, increasing the number of heads does not necessarily improve performance. Additionally, the expanding ratio (r) plays a significant role in the performance. Compared to row C, doubling r results in a $1.04\%$ increase in DSC and a $1.82\%$ drop in HD. More information regarding the technical design of the DLF module is provided in the SM.
\begin{table}[t]
    \centering
    \caption{Comparison of model parameters.}
    \vspace{-0.5em}
    \label{table:model parameters}
    \resizebox{0.9\columnwidth}{!}{%
    \begin{tabular}{cccc}
    \hline
     Model & \multicolumn{1}{l}{\# Params (M)} & \multicolumn{1}{l}{{DSC $\uparrow$}} & \multicolumn{1}{l}{HD $\downarrow$} \\ \hline
    TransUnet   & 105.28 & 77.48 &  31.69      \\
    Swin-Unet   & 27.17 & 79.13 & 21.55      \\
    LeVit-Unet   & 52.17 & 78.53 & 16.84     \\
    DeepLabv3+ (CNN)  & 59.50 & 77.63 &  39.95      \\ 
    \midrule
    \textbf{HiFormer-S}  & \textbf{23.25} & \textbf{80.29} &  \textbf{18.85}      \\ 
    \textbf{HiFormer-B}  & \textbf{25.51} & \textbf{80.39} & \textbf{14.70}      \\ 
    \textbf{HiFormer-L}  & \textbf{29.52} & \textbf{80.69} &  \textbf{19.14}     \\ \hline
    \end{tabular} %
    }
    \vspace{-1em}
\end{table}

\begin{table}[t]
    \centering
    \caption{Impact of the DLF module on the \textit{Synapse} dataset.}
    \vspace{-0.5em}
    \label{table:DLF}
    \resizebox{0.7\columnwidth}{!}{%
    \begin{tabular}{cccc}
    \hline
     Model & \multicolumn{1}{l}{DLF} & \multicolumn{1}{l}{{DSC $\uparrow$}} & \multicolumn{1}{l}{HD $\downarrow$} \\ \hline
    HiFormer-B   & \xmark     & 77.15 &  16.88    \\
    HiFormer-B   & \checkmark & 80.39 &  14.70      \\ \hline
    \end{tabular} %
    }
    \vspace{-1.5em}
\end{table}

\begin{table}[t]
    \centering
    \vspace{-0.5em}
    \caption{Comparison of different backbones for the CNN module on the \textit{Synapse} dataset. Except for the CNN module, all configurations are identical to HiFormer-B.}
    \vspace{-0.5em}
    \label{table:CNN}
    \resizebox{0.94\columnwidth}{!}{%
    \begin{tabular}{cccc}
    \hline
     Model & \multicolumn{1}{l}{\# Params (M)} & \multicolumn{1}{l}{{DSC $\uparrow$}} & \multicolumn{1}{l}{HD $\downarrow$} \\ \hline
    HiFormer+ResNet18   & 19.36 & 77.15 &  16.88      \\
    HiFormer+ResNet34   & 24.75 & 79.39 & 22.71      \\
    \textbf{HiFormer+ResNet50} & \textbf{25.51} & \textbf{80.39} & \textbf{14.70}     \\
    HiFormer+ResNet101  & 44.50 & 79.42 & 17.18      \\ 
    HiFormer+DenseNet121  & 23.92 & 78.65 &  16.18      \\ 
    HiFormer+DenseNet169  & 29.55 & 78.73 &  15.94      \\ 
    HiFormer+DenseNet201  & 35.36 & 79.08 &  21.30      \\ \hline
    \end{tabular} %
    }
    \vspace{-1.4em}
\end{table}

\noindent\textbf{Ablation on feature consistency.}
We conduct two experiments to measure and demystify the feature consistency and discuss them in detail in the SM. First, we present the feature visualization of each level before and after involving the DLF module (SM, Fig. 1-2). The second experiment proves how applying each module aids with feature consistency (SM, Table 3). Overall, the contribution of each module in providing more consistent features can be inferred from the results. 
\vspace{-1em}
\section{Discussion}
\vspace{-0.75em}
Our comprehensive experiments on different medical image segmentation datasets demonstrate the effectiveness of our proposed HiFormer model compared to CNN and transformer-based approaches. The key advances of our approach are two folds. The first rationality of its design is combining CNN and transformer both in the shallow layers of the network. Second, the skip-connection module provides feature reusability and blends CNN local features with global features provided by the transformer module. The quantitative view of the HiFormer network on five challenging datasets reveals that it can perform segmentation well, surpassing the SOTA methods in most cases. From the perspective of visual analysis, Fig. \ref{fig:synapseviz} illustrates noise-less segmentation of organs such as the Liver and Kidney, which is also consistent with quantitative benchmarks. In contrast, our model acquires failure cases in some cases (e.g., Aorta), which again agrees with numerical results. Moreover, it is perceived that the low-contrast skin images still bring great difficulties for our model. In general, HiFormer has shown the potential to learn the critical anatomical relationships represented in medical images effectively. In terms of model parameters, HiFormer is a lightweight model compared with other complex models, which impose serious problems in medical image segmentation.
\begin{table}[t]
\centering
\vspace{-0.5em}
    \caption{Ablation study for the DLF module with different parameters on the \textit{Synapse} dataset. For a fair comparison, ResNet-50 has been used as the CNN module in all the configurations, and $r$ denotes the MLP expanding ratio used in the transformer block of the DLF module.}
    \vspace{-0.5em}
    \label{table:DLF Configs}
    \begin{adjustbox}{max width=\linewidth}
    \begin{tabular}{c|ccccccl|cc|c} 
    \toprule
    \multirow{2}{*}{Model} & \multicolumn{2}{c}{Dimension} &  &  &  & \multicolumn{2}{c|}{\# Heads} &  &  & Params \\
     & $\mathrm{P^s}$ &  $\mathrm{P^l}$ & S & L & r & $\mathrm{P^s}$ &  $\mathrm{P^l}$ &  DSC $\uparrow$ & HD $\downarrow$ & (M) \\ 
    \midrule
    \rowcolor[HTML]{C8FFFD}
    HiFormer-B & 384 & 96 & 2 & 1 & 2 & 6 & 6 & \textbf{80.39} & \textbf{14.70} & 25.51 \\ 
    \midrule
    A & 384 & 96 & 2 & 1 & 2 & \textbf{\textcolor{magenta}{12}} & 6 & 79.00 & 15.81 & 25.51 \\
    B & 384 & 96 & 2 & 1 & 2 & \textbf{\textcolor{magenta}{3}} & \textbf{\textcolor{magenta}{3}} & 77.95 & 19.11 & 25.51 
    \\
    C & 384 & 96 & 2 & 1 & \textbf{\textcolor{magenta}{1}} & 6 & 6 & 79.35 & 16.52 & 24.90 
    \\
    D & 384 & 96 & 2 & 1 & \textbf{\textcolor{magenta}{3}} & 6 & 6 & 79.22 & 17.96 & 26.12 
    \\
    E & 384 & 96 & \textbf{\textcolor{magenta}{1}} & 1 & 2 & 6 & 6 & 79.48 & 20.15 & 24.33
    \\
    F & 384 & 96 & 2 & \textbf{\textcolor{magenta}{2}} & 2 & 6 & 6 & 78.86 & 19.75 & 25.59 
    \\
    
    \bottomrule
    \end{tabular}
    \end{adjustbox}
    \vspace{-1.5em}
\end{table}

\vspace{-0.75em}
\section{Conclusions}
\vspace{-0.75em}
In this paper, we introduce HiFormer, a novel hybrid CNN-transformer-based method for medical image segmentation. Specifically, we combine the global features obtained from a Swin Transformer module with local representations of a CNN-based encoder. Then, using a DLF module, we attain a finer fusion of features derived from the aforementioned representations. We achieve superior performance over CNN-based, vanilla transformer-based, and hybrid models indicating that our methodology secures the balance in keeping the details of low-level features and modeling the long-range interactions.

{\small
\bibliographystyle{ieee_fullname}
\bibliography{egbib}
}

\newpage
\thispagestyle{empty}
\appendix
\newpage

%
%
%
%
%

\section*{Supplementary Material}
This supplementary material contains the following additional information. We expand Table 6 in the paper and provide more experiments regarding the impact of the proposed DLF module (see Table \ref{table:DLFSkin}-\ref{table:DLFSegPc}). Moreover, we provide additional information regarding the optimality of DLF module and the intuitions behind fusion of features in different levels and feature consistency.

\section{Impact and justification of the DLF Module}

\subsection{Model Design Motivation}

The deficiency of transformers in capturing local features, lack of data in the medical domain, and proven usage of CNN-produced features as an input to transformers and their success in vision tasks \cite{chen2021transunet} led us to use a rich CNN backbone before the transformer. Subsequently, we used a successive Swin Transformer to capture multi-scale global dependencies. With respect to the deformation of body organs and tissues and diverse sizes and scales of neighboring organs, we proposed a representative token by applying GAP to form a representative token or messenger token-like \cite{fang2022msg,farooq2021global} to exchange information between scales to prevent the globality bias to a specific region and implicitly remembers its previous stage attention regions. Our expansion of ablation study for DLF module presence is depicted in Table \ref{table:DLFSkin} and Table \ref{table:DLFSegPc}. 

In addition, the inspiration behind presenting variable HiFormer designs is to develop a general model with different scales and exhibit the stability of the model. Considering the accuracy-speed trade-off, one can exploit the S, L, or B HiFormer and investigate the whole network performance.

\subsection{Hyper-parameter Optimization}

\noindent\textbf{S, L, r:} We have considered the effect of network deepening and model scaling on the network performance similar to \cite{chen2021transunet,cao2021swin}. In our experiments, we deduced that increasing the $(S, L,r)$ pairs would lead to a substantial computational cost which is in contradiction to our main contribution of designing a stable model with low parameters. Specifically, we hypothesized that considering $S, L,r >3$ (see row D in table 8) can result in the model overparameterization along with the extraction of redundant features. Hence, we adopted $(S, L,r)<=3$. 

\textbf{Number of heads.} Considering the number of heads of the transformer, we performed cross-validation using the synapse dataset and attained 6 heads as the ideal choice. For the sake of demonstrating the effect of the number of transformer heads, two more configurations besides 6, its half, and double (3 and 12) were considered in our ablation study.


\begin{table}[h]
	\centering
	\caption{Impact of the DLF module on the skin lesion segmentation datasets.}
	\label{table:DLFSkin}
	\resizebox{1\columnwidth}{!}{%
		\begin{tabular}{cccccc}
			\hline
			\textbf{Model} &\textbf{DLF} & \textbf{DSC} & \textbf{SE} & \textbf{SP} & \textbf{ACC} \\ \hline
			\multicolumn{6}{c}{\textit{ISIC 2017}} \\ \hline
			HiFormer-B   & \xmark   & 0.9167 &  0.8814 & \textbf{0.9895} & 0.9678   \\
			\rowcolor[HTML]{d2dfde}
			HiFormer-B   & \checkmark & \textbf{0.9253} & \textbf{0.9155} & 0.9840 & \textbf{0.9702}\\ \hline
			\multicolumn{6}{c}{\textit{ISIC 2018}}\\ \hline
			HiFormer-B   & \xmark   & 0.8986 &  0.8559 & \textbf{0.9870} & 0.9595   \\
			\rowcolor[HTML]{d2dfde}
			HiFormer-B   & \checkmark & \textbf{0.9102} &  \textbf{0.9119} & 0.9755 & \textbf{0.9621}    \\ \hline
			\multicolumn{6}{c}{\textit{PH$^2$}}\\ \hline
			HiFormer-B   & \xmark   & 0.9321 &  0.9016 & \textbf{0.9848} & 0.9586   \\
			\rowcolor[HTML]{d2dfde}
			HiFormer-B   & \checkmark & \textbf{0.9460} &  \textbf{0.9420} & 0.9772 & \textbf{0.9661}    \\ \hline
		\end{tabular} %
	}
\end{table}

\begin{table}[h]
	\centering
	\caption{Impact of the DLF module on the \textit{SegPC} dataset.}
	\label{table:DLFSegPc}
	\resizebox{0.65\columnwidth}{!}{%
		\begin{tabular}{ccc}
			\hline
			\textbf{Model} & \textbf{DLF} & \textbf{mIoU} \\ \hline
			HiFormer-B   & \xmark     & 0.9317     \\
			\rowcolor[HTML]{d2dfde}
			HiFormer-B   & \checkmark & \textbf{0.9406}      \\ \hline
		\end{tabular} %
	}
	\vspace{-1em}
\end{table}

\section{Clarification on CNN backbones}
\noindent Our study considered different backbones typically used in the literature \cite{chen2021transunet}, such as ResNet and DenseNet. Benefiting from the skip-connection criteria, the ResNet architecture can facilitate multi-level representation. Although a DensNet or ResNet with more layers might bring a stronger representation, it can be a more high-level representation so that its amalgamation with the transformer in subsequent layers would prevent the transformer from extracting better features. Moreover, the features attained from a ResNet with 50 layers can be considered as more general ones compared to 18, 34 layered shallow ResNets aiding the transformer in more optimal performance.


\section{Feature Consistency}
\begin{figure*}[htb]
	\centering
	\resizebox{0.98\textwidth}{!}{
		\begin{tabular}{@{} *{6}c @{}}
			\includegraphics[width=0.18\textwidth]{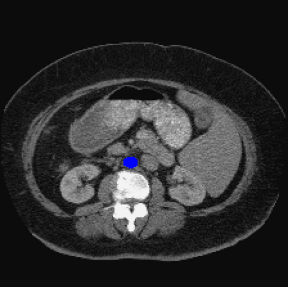} &
			\includegraphics[width=0.18\textwidth]{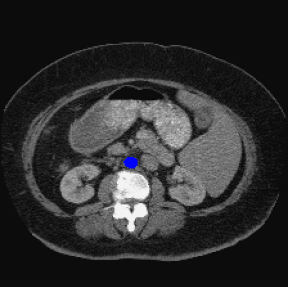} &
			\includegraphics[width=0.18\textwidth]{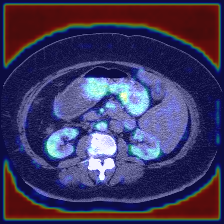} &
			\includegraphics[width=0.18\textwidth]{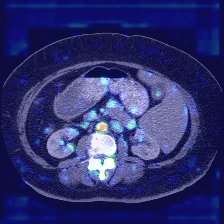} &
			\includegraphics[width=0.18\textwidth]{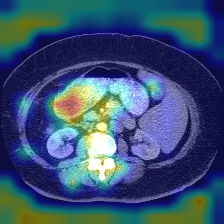} &
			\includegraphics[width=0.18\textwidth]{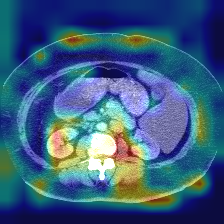} 
			\\
			\includegraphics[width=0.18\textwidth]{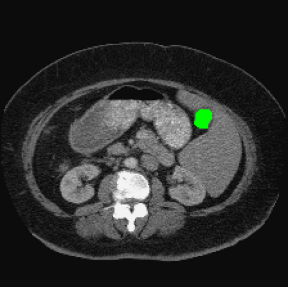} &
			\includegraphics[width=0.18\textwidth]{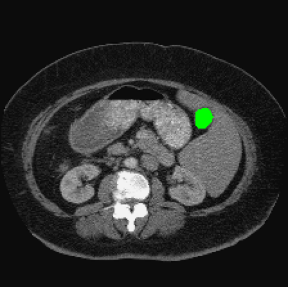} &
			\includegraphics[width=0.18\textwidth]{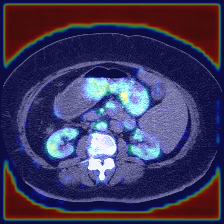} &
			\includegraphics[width=0.18\textwidth]{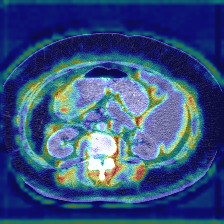} &
			\includegraphics[width=0.18\textwidth]{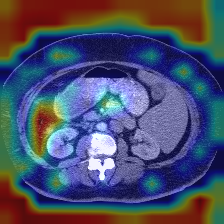} &
			\includegraphics[width=0.18\textwidth]{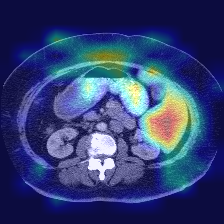} 
			\\
			\includegraphics[width=0.18\textwidth]{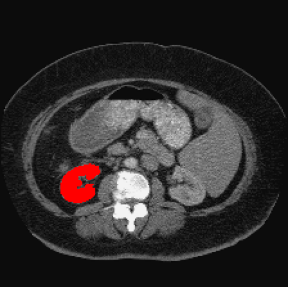} &
			\includegraphics[width=0.18\textwidth]{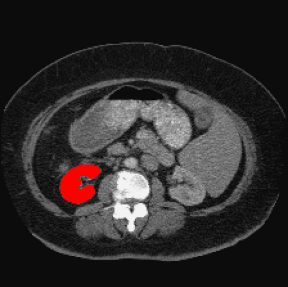} &
			\includegraphics[width=0.18\textwidth]{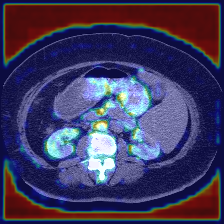} &
			\includegraphics[width=0.18\textwidth]{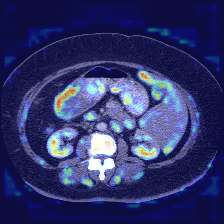} &
			\includegraphics[width=0.18\textwidth]{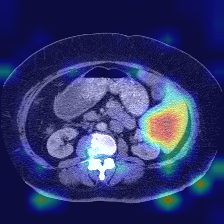} &
			\includegraphics[width=0.18\textwidth]{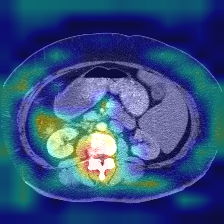} 
			\\
			\includegraphics[width=0.18\textwidth]{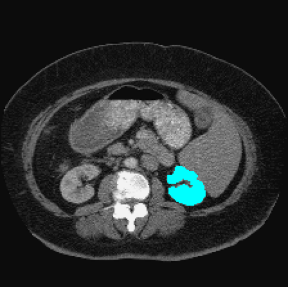} &
			\includegraphics[width=0.18\textwidth]{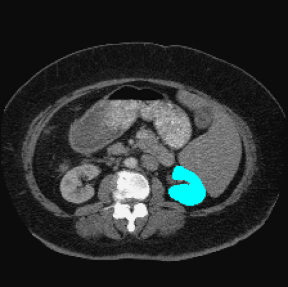} &
			\includegraphics[width=0.18\textwidth]{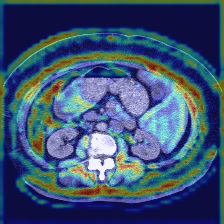} &
			\includegraphics[width=0.18\textwidth]{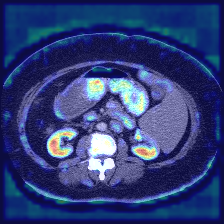} &
			\includegraphics[width=0.18\textwidth]{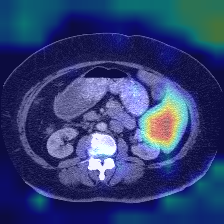} &
			\includegraphics[width=0.18\textwidth]{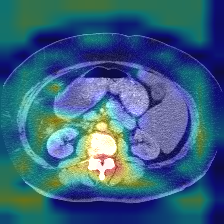} 
			\\
			\includegraphics[width=0.18\textwidth]{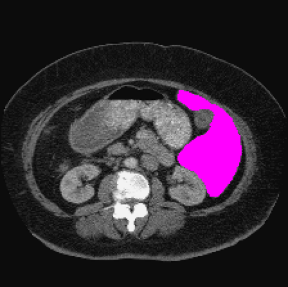} &
			\includegraphics[width=0.18\textwidth]{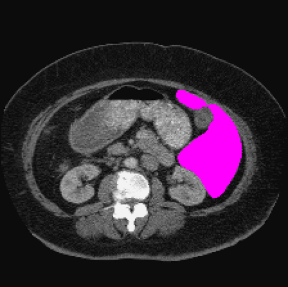} &
			\includegraphics[width=0.18\textwidth]{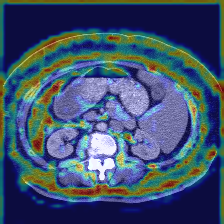} &
			\includegraphics[width=0.18\textwidth]{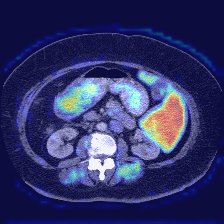} &
			\includegraphics[width=0.18\textwidth]{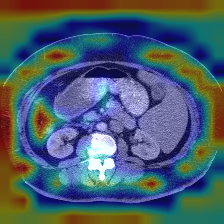} &
			\includegraphics[width=0.18\textwidth]{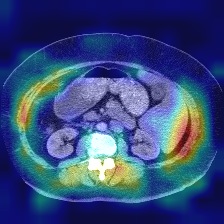} 
			\\
			\includegraphics[width=0.18\textwidth]{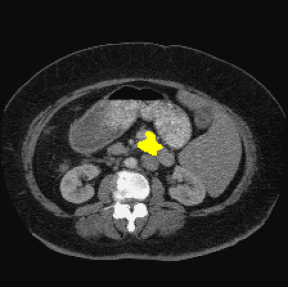} &
			\includegraphics[width=0.18\textwidth]{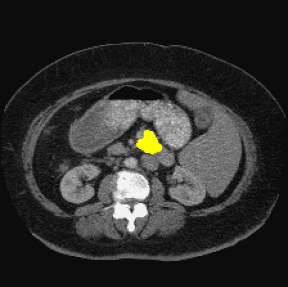} &
			\includegraphics[width=0.18\textwidth]{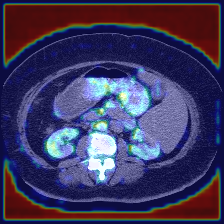} &
			\includegraphics[width=0.18\textwidth]{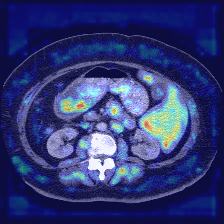} &
			\includegraphics[width=0.18\textwidth]{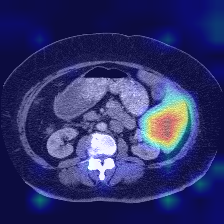} &
			\includegraphics[width=0.18\textwidth]{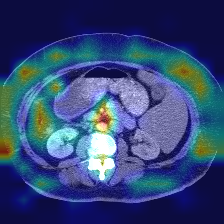} 
			\\
			{ \large (a) Ground Truth} & {\large (b) HiFormer-B} & {\large (c) $P^l$ Before DLF} & {\large (d) $P^l$ After DLF} & {\large (e) $P^s$ Before DLF} & {\large (f) $P^s$ After DLF}
		\end{tabular}
	}
	\caption{Feature visualization of HiFormer-B using Grad-CAM \cite{selvaraju2017grad}.}
	\label{fig:test1}
\end{figure*}

\begin{figure*}[htb]
	\centering
	\resizebox{0.98\textwidth}{!}{
		\begin{tabular}{@{} *{6}c @{}}
			\includegraphics[width=0.18\textwidth]{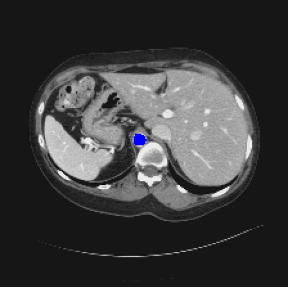} &
			\includegraphics[width=0.18\textwidth]{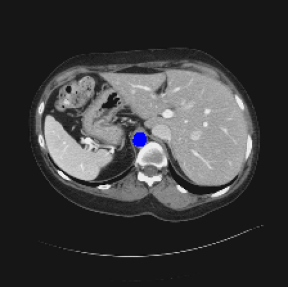} &
			\includegraphics[width=0.18\textwidth]{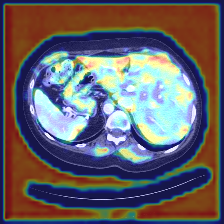} &
			\includegraphics[width=0.18\textwidth]{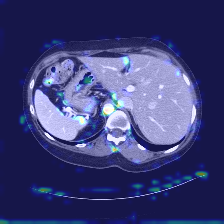} &
			\includegraphics[width=0.18\textwidth]{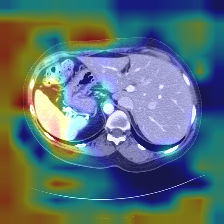} &
			\includegraphics[width=0.18\textwidth]{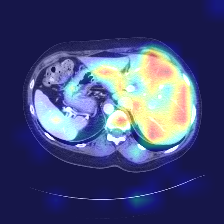} 
			\\
			\includegraphics[width=0.18\textwidth]{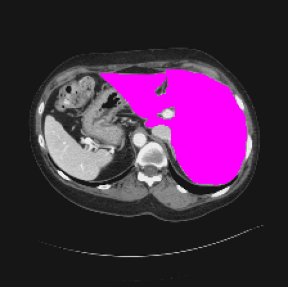} &
			\includegraphics[width=0.18\textwidth]{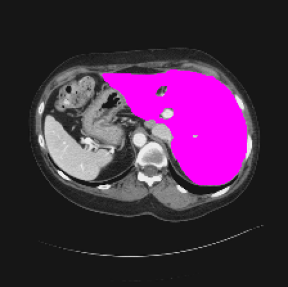} &
			\includegraphics[width=0.18\textwidth]{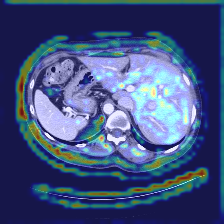} &
			\includegraphics[width=0.18\textwidth]{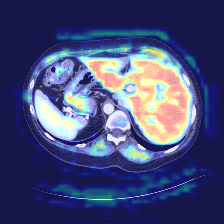} &
			\includegraphics[width=0.18\textwidth]{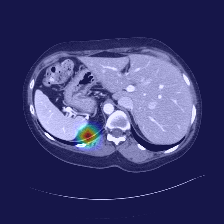} &
			\includegraphics[width=0.18\textwidth]{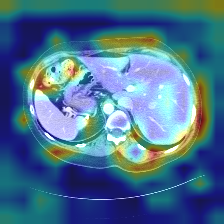} 
			\\
			\includegraphics[width=0.18\textwidth]{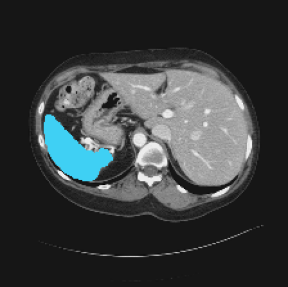} &
			\includegraphics[width=0.18\textwidth]{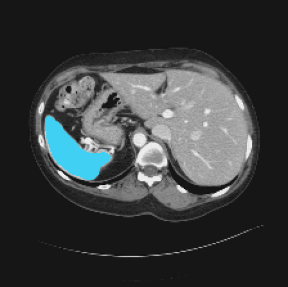} &
			\includegraphics[width=0.18\textwidth]{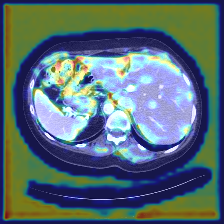} &
			\includegraphics[width=0.18\textwidth]{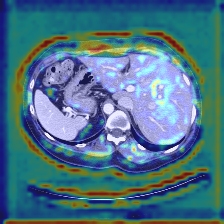} &
			\includegraphics[width=0.18\textwidth]{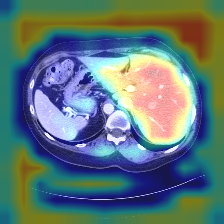} &
			\includegraphics[width=0.18\textwidth]{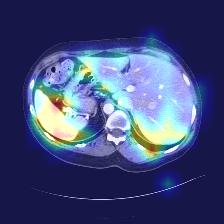} 
			\\
			\includegraphics[width=0.18\textwidth]{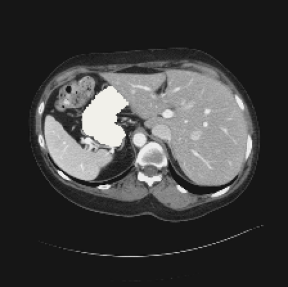} &
			\includegraphics[width=0.18\textwidth]{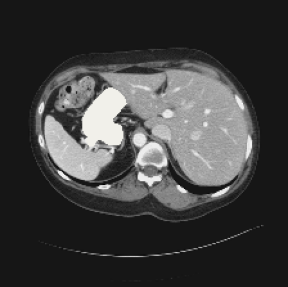} &
			\includegraphics[width=0.18\textwidth]{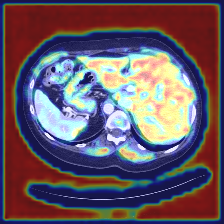} &
			\includegraphics[width=0.18\textwidth]{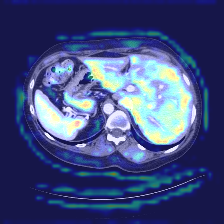} &
			\includegraphics[width=0.18\textwidth]{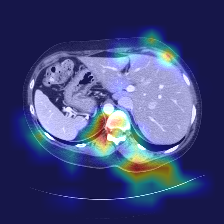} &
			\includegraphics[width=0.18\textwidth]{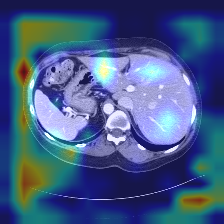} 
			\\
			{ \large (a) Ground Truth} & {\large (b) HiFormer-B} & {\large (c) $P^l$ Before DLF} & {\large (d) $P^l$ After DLF} & {\large (e) $P^s$ Before DLF} & {\large (f) $P^s$ After DLF}
		\end{tabular}
	}
	\caption{Feature visualization of HiFormer-B using Grad-CAM \cite{selvaraju2017grad}.}
	\label{fig:test2}
\end{figure*}

\noindent We provide two experiments to demystify the feature consistency.
First, we present the feature visualization of each level before and after involving the DLF module (see Fig. \ref{fig:test1}-\ref{fig:test2}). As illustrated, before the DLF module is applied, the attention location is more diffused, therefore the organ is not clearly emphasized. However, after applying the DLF module, attention is drawn to the desired organ and is more highlighted surrounding the organ, demonstrating that the DLF module makes features more consistent. Furthermore, both levels serve a complimentary function, with the larger level providing fine-grained features and the smaller level attempting to give extra information. As a result, both levels are required for the model to function effectively.

In the second experiment, we take the HiFormer-B and remove modules in a hierarchical order to observe how the features become consistent. As shown in Table \ref{table:feature}, using only ResNet50 as the CNN module and dismissing others achieves a 77.40 dice score and 26.71 HD. Having involved the Swin Transformer, HD witnesses a 9.93 drop, indicating that our predictions become closer to their corresponding labels or more similar. Subsequently, applying the DLF module not only increases the dice score but also decreases HD, exhibiting that the module dramatically assists in making the features consistent.

\begin{table}[h]
	\centering
	\caption{Impact of each module in HiFormer-B.}
	\label{table:feature}
	\resizebox{1\columnwidth}{!}{%
		\begin{tabular}{cccccc}
			\hline
			\textbf{Model} & \textbf{CNN} & \textbf{transformer} & \textbf{DLF} & \textbf{DSC} & \textbf{HD}
			\\ \hline
			HiFormer-B   & \checkmark     & \xmark & \xmark & 77.40 & 26.71    \\
			HiFormer-B   & \checkmark & \checkmark & \xmark & 77.15 & 16.88      \\ 
			\rowcolor[HTML]{d2dfde}
			HiFormer-B   & \checkmark & \checkmark & \checkmark & \textbf{80.39} & \textbf{14.70}      \\ \hline
		\end{tabular} %
	}
	\vspace{-1em}
\end{table}

\end{document}